\documentclass[conference]{IEEEtran}
\IEEEoverridecommandlockouts
\usepackage{cite}
\usepackage{comment}
\usepackage{amsmath,amssymb,amsfonts}
\usepackage{algorithmic}
\usepackage{graphicx}
\usepackage{textcomp}
\usepackage{xcolor}
\usepackage[normalem]{ulem}
\usepackage{multicol}
\usepackage{algorithm}

\def\BibTeX{{\rm B\kern-.05em{\sc i\kern-.025em b}\kern-.08em
    T\kern-.1667em\lower.7ex\hbox{E}\kern-.125emX}}
\usepackage[symbol]{footmisc}

\newcommand{\blue}[1]{#1}

\usepackage{tikz}

\begin{document}

\title{Efficient Synaptic Delay Implementation in Digital Event-Driven AI Accelerators}

\author{
Roy Meijer\textsuperscript{b},
Paul Detterer\textsuperscript{c},
Amirreza Yousefzadeh\textsuperscript{c, d},
Alberto Patiño-Saucedo\textsuperscript{a},

Guanghzi Tang\textsuperscript{c},\\
Kanishkan Vadivel\textsuperscript{c},
Yinfu Xu\textsuperscript{c},

Manil-Dev Gomony\textsuperscript{b},
Federico Corradi\textsuperscript{b},  \\
Bernabé Linares-Barranco\textsuperscript{a},
and Manolis Sifalakis\textsuperscript{c} \\
\textsuperscript{a} Instituto de Microelectrónica de Sevilla, CSIC/Universidad de Sevilla, Seville, Spain,\\
\textsuperscript{b} Eindhoven University of Technology, Dept. Electrical Engineering, Eindhoven, Netherlands,\\
\textsuperscript{c} IMEC-Netherlands, Eindhoven, Netherlands, \\
\textsuperscript{d} University of Twente, Twente, Netherlands\\
Corresponding author email: manolis.sifalakis@imec.nl\\
}

\maketitle

\begin{abstract}

Synaptic delay parameterization of neural network models have remained largely unexplored but recent literature has been showing promising results, suggesting the delay parameterized models are simpler, smaller, sparser, and thus more energy efficient than similar performing (e.g. task accuracy) non-delay parameterized ones. We introduce Shared Circular Delay Queue (SCDQ), a novel hardware structure for supporting synaptic delays on digital neuromorphic accelerators. Our analysis and hardware results show that it scales better in terms of memory, than current commonly used approaches~\cite{akopyan2015truenorth,davies2018loihi,khan2008spinnaker}, and is more amortizable to algorithm-hardware co-optimizations, where in fact, memory scaling is modulated by model sparsity and not merely network size. Next to memory we also report performance on latency area and energy per inference.

\end{abstract}

\begin{IEEEkeywords}
Spiking Neural Networks, Synaptic Delays, Hardware Acceleration
\end{IEEEkeywords}

\section{Introduction}
\label{sec:intro}

In biological neural networks synaptic delays are known to be subject to learning and play an important role in the propagation and processing of information~\cite{stoelzel2017}.
Hence, configurable delays are a basic feature offered in many neuromorphic neural-network accelerators~\cite{akopyan2015truenorth, khan2008spinnaker, davies2018loihi, starzyk2020}.
However, up until recently synaptic delays have rarely been used in practice, because (i) it has been unclear how to parameterize and optimize them in models and (ii) because their use has been assumed expensive and in terms of memory and complicated in digital accelerators.

Interestingly, in the last couple of years, progress has been made with effectively training models parameterized with synaptic delays~\cite{patino2023empirical, zhang2020, hammouamri2023learning, wang2019, sun2023adaptive}, which shows that these models are more compact, and have simpler structure~\cite{patino2023empirical, sun2023adaptive}, while achieving comparable or higher performance~\cite{patino2023empirical, hammouamri2023learning}, than their counterparts with memoryless synapses. 
They have also been observed to exhibit sparser activation patterns~\cite{patino2023empirical}.
This raises the question whether the compactness and sparseness of the model outweigh the speculated memory, energy, and latency costs of the synaptic delay implementation in hardware.
The study in~\cite{patino2023empirical} confirmed that this appears to be indeed the case when comparing the memory and energy estimates of the models with and without synaptic delays.
These estimates were obtained by considering synaptic activity in recurrent and feed-forward models trained with and without synaptic delays, and when using common hardware structures for synaptic delays that are used in Loihi~\cite{davies2018loihi} and TrueNorth~\cite{akopyan2015truenorth}.

The questions that emerged and motivated this work, concern: (i) the scalability of delay structures as model size and complexity grow;
and (ii) the efficiency of incorporating synaptic delay structures in hardware, considering factors such as memory, IC area and inference latency, which are relevant in edge AI applications.

In this paper we introduce a new synaptic delay structure and its implementation in a digital circuit design for multicore neuromorphic accelerators. It is based on a circular shared-queue and its operation is leveraged by a simple extension of the Address Event Representation (AER) packet format~\cite{gillespie1993}, commonly used in neuromorphic systems for communicating events. Compared to existing approaches~\cite{akopyan2015truenorth, khan2008spinnaker, davies2018loihi}, our approach is more memory and area efficient because its overhead does not increase with the size of the network models, but rather with the neural network's activation density. This activation density can be minimized during training, thus, allowing for a hardware-algorithm co-optimization. Importantly, the introduction of synaptic delays has minimal impact on power, area, and energy requirements of the accelerator.

We have evaluated our hardware implementation on the Seneca neuromorphic platform~\cite{tang2023Seneca}. We report on inference fidelity, hardware metrics (energy per inference, latency, area), and some comparisons with other reference digital neuromorphic systems that implement delays~\cite{akopyan2015truenorth, davies2018loihi}.

\section{Background and Related Work}
\label{sec:background}

In Spiking Neural Networks (SNNs), information is conveyed through pulses that travel between the presynaptic and postynaptic neurons. These pulses encode a specific weight associated with the synaptic connection, and if there are synaptic delays incorporated into the model, these pulses require some time to propagate from the presynaptic neuron to the postynaptic one.

\begin{figure}
    \centering
        \centering
        \includegraphics[width=0.18\textwidth, trim=20 20 20 20, clip]{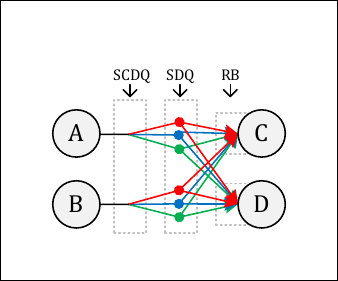}
    \hfill
    \caption{\small Example of two SNN delay models with two presynaptic neurons, two postynaptic neurons, and three levels of delay.}
    \label{fig:delay_model}
\end{figure}

In~\cite{patino2023empirical}, Patino et al. introduced a simple (i.e. hardware-friendly) framework for parameterizing and training models with synaptic delays, whereby for each discrete delay level $D$, there is an additional synapse between each presynaptic neuron and postynaptic neuron pair, each with a unique trainable weight. Spikes from presynaptic neurons are received by postynaptic neurons over $D$ discrete time steps, with varying weights. The number of delay levels determines how many times a postynaptic neuron receives the same repeated spike from a presynaptic neuron.
In this case, delay parameterization involves extending the weight matrix structure from 2D to 3D (2D weight matrix for each delay value). It has dimensions $D \times I \times J$, where $D$ is the delay resolution (that is, the number of distinct discrete values), $I$ is the number of presynaptic neurons, and $J$ is the number of postynaptic neurons.
Figure \ref{fig:delay_model} illustrates such a delay-parameterized network with $D = 3$. The weight matrix has dimensions $3 \times 2 \times 2$. Dots in the middle of connections between neuros symbolize synapses, and on their left and right are, respectively, axons from presynaptic neurons and dendrites to postynaptic neurons. Color coding represents the delay levels. Each delay axon combined with a weighted dendritic branch defines a \emph{delay synapse} making a total of $3 \times 2 \times 2 = 12$ weights. Pruning an axon removes a set of delay synapses towards postynaptic layer neurons, and pruning a dendritic branch alone removes an individual synaptic connection between a pre-post neuron pair.

In most neuromorpchic architectures, delays are implemented as FIFO queues or delay lines placed either at every neuron or shared among synaptic connections.
    
For example, in~\cite{davies2018loihi, khan2008spinnaker,morrison2005advancing} a \emph{Ring Buffer} is a circular FIFO queue placed at each postsynaptic neuron where dendritic weights accumulate. Each \emph{slot} of the Ring Buffer accumulates weights for a different timestep. At the end of a timestep, the slot corresponding to the current timestep is added to the membrane potential of the corresponding postsynaptic neuron. Subsequently, the Ring Buffer shifts to the next timestep, and the slot from the current timestep becomes the slot for the maximum delay. The number of delay levels is constrained by the size (number of slots) of the Ring Buffer. With only one Ring Buffer per postynaptic neuron, the total memory overhead is the product of the number of postynaptic neurons $J$ and the number of delay levels $D$, scaling with $\mathcal{O}(J \cdot D)$ \cite{patino2023empirical}.

On the other hand, a \emph{Shared Delay Queue}~\cite{akopyan2015truenorth} is a structure comprising multiple FIFOs in a linear cascade arrangement shared across the synapses (presynaptic and postynaptic neuron pairs of two layers). The number of FIFOs in the cascade equals the maximum number of discrete delay levels $D$. Every event triggered at a presynaptic neuron from an axon with a given delay, enters the Shared Delay Queue at the FIFO corresponding to the number of timesteps it needs to wait before delivery. At the end of a timestep, all events in every FIFO shift to the next FIFO in the cascade belonging to the next relative timestep, and all events in the FIFO belonging to the current timestep are transmitted to the postynaptic neurons. The total memory usage depends on the minimum activation sparsity of the network (which may be biased through model training). Assuming a minimum activation sparsity of $\alpha$, where $\alpha = 1$ implies that every presynaptic neuron is activated and $\alpha = 0$ implies that none are activated, the total memory overhead is calculated as $\alpha \cdot I \cdot \left(\sum_{d=1}^{D}(D-d)\right) = \frac{1}{2}\cdot \alpha \cdot I \cdot (D^2 + D)$, where $I$ is the number of presynaptic neurons and $D$ is the number of delay levels. This memory overhead scales as $\mathcal{O}(\alpha \cdot I \cdot D^2)$. In this synaptic delay model only axonal delays are supported and therefore an event enters and exist the Shared Delay Queue only once.

\section{Synaptic Delay Architecture}
\label{sec:method}

\subsection{Shared Circular Delay Queue Structure}
\label{subsec:model}

\begin{figure}
\centering
\includegraphics[width=0.48\textwidth,trim=15 15 15 15,clip]{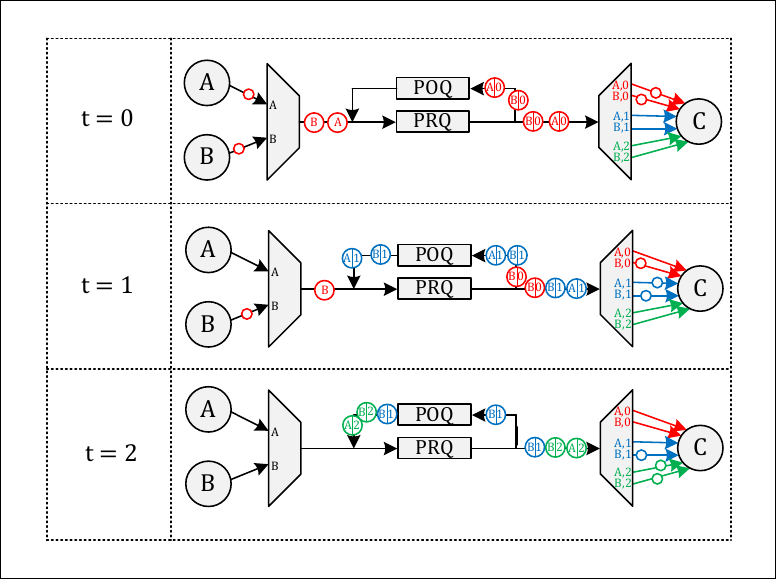}
\caption{\small An example of the event flow over three timesteps in a two-layer SNN with synaptic delays. The maximum delay is 2, and the Shared Circular Delay Queue is positioned between the two layers. In timestep $t=0$, neurons $A$ and $B$ spike, and neuron $C$ receives spikes from neurons $A$ and $B$ with a delay value of 0. In timestep $t=1$, neuron $B$ spikes, and neuron $C$ receives a spike from neuron $A$ with a delay value of 0, from neuron $B$ with a delay value of 0, and from neuron $B$ with a delay value of 1. In timestep $t=2$, no neuron spikes, and neuron $C$ receives a spike from neuron $A$ with a delay value of 2, from neuron $B$ with a delay value of 2, and from neuron $B$ with a delay value of 1.}
\label{fig:example_3_timesteps}
\end{figure}

The proposed synaptic delay structure draws inspiration from the shared queue model in ~\cite{akopyan2015truenorth} and incorporates the concept of ``double-buffering''. Instead of a linear chain of FIFOs (one per delay level), it simplifies the design to only two FIFOs: the \emph{pre-processing queue} (PRQ) and the \emph{post-processing queue} (POQ). These two FIFOs are interconnected in a circular arrangement (see Figure \ref{fig:example_3_timesteps}), which is why we refer to this configuration as the ``Shared Circular Delay Queue'' (SCDQ) in this paper.

There is an SCDQ module at every compute core of a neural network accelerator. Conceptually it is positioned between a presynaptic and a postsynaptic layer, with the input of the SCDQ queue facing the presynaptic layer and the output of the SCDQ facing the postsynaptic layer, and is shared across all synapses between neuron pairs. If multiple layers can be mapped in the same compute core then an SCDQ can also be shared across layers. As a result the memory complexity is not a function of the number of layers in a model, but rather a function of the number of compute cores in use (and the aggregate activation traffic).

When an event packet (AER) is received at the SCDQ input, a delay counter will be added to it, which counts the maximum number of algorithm timesteps by which the event should be delayed. The event then traverses the PRQ and arrives at the output if delivery to some postsynaptic neuron is expected at the current algorithm timestep. It will (\emph{also}) be pushed to the POQ if it is expected to be delayed for delivery at a future timestep. At the POQ the delay counter of events is decremented for one timestep. At some point, a special event packet signaling the \emph{end of the algorithmic timestep} intercepted at the output, will trigger a buffer-swap between POQ and PRQ. When this happens, the events previously held in POQ will be propagated through PRQ towards the output (if their delay counter has reached a delay value due for delivery in this new timestep) and/or will be pushed back again to the POQ for further delay and delivery in a future timestep. When the delay counter of an event expires, this event will no longer be pushed to the POQ and will only be forwarded to the output. Figure \ref{fig:example_3_timesteps} illustrates an example of the flow of events through the SCDQ across two layers of an SNN network with delays in three timesteps.

It is important to note that by contrast to its predecessor~\cite{akopyan2015truenorth}, SCDQ makes it possible for an event to ``orbit'' and exit the queue at several timepoints. This way it can support event delays not only per axon but also per axon-dendrite pair (i.e. per individual synapse), which we will discuss further in section\ref{sec:results}.

SCDC consists of six functional submodules shown in the block diagram of Figure \ref{fig:delay_ip_hw_accelerator}: the \emph{write controller} and \emph{read controller} which get triggered by external transactions, the \emph{PRQ} and \emph{POQ}, the \emph{pruning filter}, and \emph{memory}. The pruning filter and memory will be discussed in more detail in sections \ref{subsec:pruning} and \ref{subsec:area} respectively. The \emph{write controller} is responsible for reading the events from the input and writing them to the PRQ and adding to them an initialized delay counter value. Additionally, it is responsible for swapping the PRQ and POQ buffers when a special \emph{end-of-timestep} (EOT) event is intercepted at the output. The \emph{read controller} is responsible for reading events from the PRQ and writing them to the output and the POQ while decreasing the delay counter. It will also signal the write controller to start the EOT sequence when there is an EOT event at the output. The PRQ and POQ are FIFO abstractions around memory buffers available in the memory module, and thereby implement the actual data-path through the SCDQ.


\subsection{Zero-Skipping Delay-Forwarding}
\label{subsec:pruning}


One observes in Figure \ref{fig:delay_model} that the network structure can become quite intricate due to the presence of numerous synapses. The number of synapses increases linearly with the maximum number of delay steps, $D$, which we aim to accommodate. In practical scenarios, weight training, quantization, and the potential use of suitable regularization techniques, can efficiently remove a substantial number of synapses~\cite{patino2023empirical}. This results in a much sparser or pruned model that is well suited for effective inference in applications.

In a zero-skipping capable accelerator (typically most neuromorphic, but also several conventional accelerators) a zero weight or activation will skip loading weights from memory and carrying out the partial MAC, thereby saving energy and likely improving latency.
When accommodating synaptic delays this decision is to be taken after an event exits the shared delay queue and is about to be received by the postsynaptic neurons. 
This can unnecessarily occupy FIFO memory, consume power for repeated lookups in weight-memory, and affect latency in SCDQ; if some or all of the postsynaptic neuron connection have zero weights.

\begin{figure}[h]
    \centering
    \begin{minipage}{0.3\textwidth}
        \includegraphics[width=\linewidth,trim=15 15 15 15,clip]{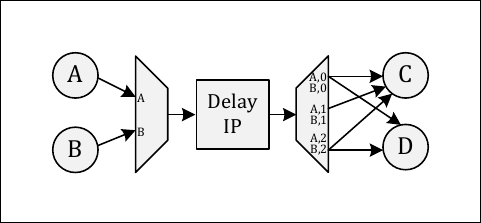}
    \end{minipage}\hfill
    \begin{minipage}{0.15\textwidth}
        \[
        WVU =
        \begin{bmatrix}
        1 & 1 & 0\\
        0 & 0 & 1
        \end{bmatrix}
        \]
    \end{minipage}
    \caption{\small Example of $WVU$ for a network where delayed axons $A,2$, $B,0$ and $B,1$ are skipped.}
    \label{fig:wvu_example}
\end{figure}

To reduce this overhead a \emph{zero-skiping delay-forwarding} capability is implemented in the \emph{pruning filter} of SCDQ.
A binary matrix called $WVU$ (which stands for ``weight value useful'') is maintained locally, where every element represents a delayed axon.
$WVU$ has two dimensions: dimension $I$ representing every presynaptic neuron, and dimension $D$ representing every delay level.
If $WVU_{i,d} = 1$, this means that for the current delay value $d$ and the presynaptic neuron address $i$, at least one of the weights is a non-zero value. 
If $WVU_{i,d} = 0$, this means that for all $i$ and $d$, the weights are 0
(which implies that event delivery can be skipped in this timestep).
Using this method, the SCDQ can determine for every event in the PRQ if it is useful for the postsynaptic layer (or if it should be ignored).
Figure \ref{fig:wvu_example} shows an example of a network with some pruned delayed axons and the corresponding  $WVU$ matrix.

To determine when an event should be removed from the delay queue (i.e. every next timestep of the event, $WVU_{i,d}$ will be 0), a row of $WVU$, which corresponds to some presynaptic neuron $i$, can be treated as a binary number. If we count the number of leading zeros ($clz(x)$) of this binary number, we know for which delay counter value $d$ the event does not have to move from the PRQ to the POQ anymore. For example, in Figure \ref{fig:wvu_example}, $clz(WVU_{A}) = 1$ and $clz(WVU_{B}) = 0$, which means events from neuron $A$ have to stay for 2 timesteps (can be removed at delay counter 1) and events from neuron $B$ do not have to stay for 3 timesteps (can be removed at delay counter 0).

\subsection{Hardware IP in Seneca Neuromorphic processor}

To evaluate the SCDQ structure, we implemented a hardware IP block (Delay IP) and integrated it in the Seneca~\cite{tang2023Seneca} neuromorphic processor. We also implemented a software version of it that could be run on the RISC-V controller of a Seneca core (NCC).

Seneca is a fully digital time-multiplexed multicore accelerator similar to Loihi~\cite{davies2018loihi}, SpiNNaker~\cite{khan2008spinnaker} and TrueNorth~\cite{akopyan2015truenorth}. While SpiNNaker and Loihi use ring buffers to support synaptic delays, TrueNorth and Seneca (through the Delay IP of SCDQ) use variations of the general shared queue structure explained in section~\ref{sec:background}. 

\begin{figure}
\centering
\includegraphics[width=0.48\textwidth,trim=22 18 22 23,clip]{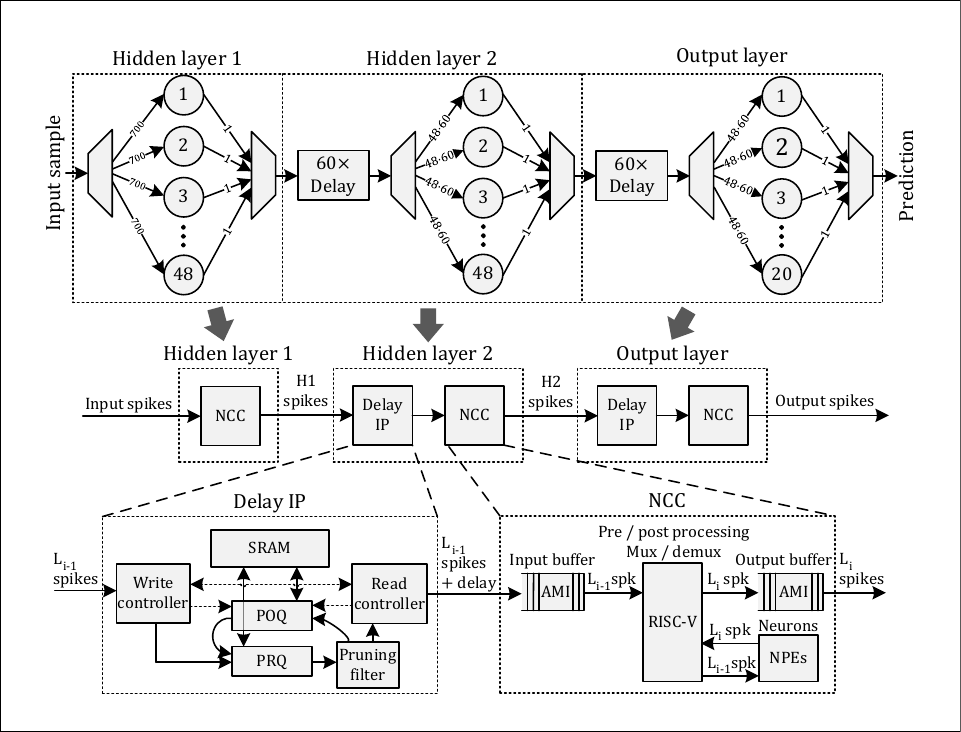}
\caption{\small A mapping on the Seneca platform of a three-layer SNN with 700 wide input samples, 48 neurons in both hidden layers, 20 neurons in the output layer, and 60 delays between the hidden layers}
\label{fig:multi_layer_network_everything}
\end{figure}

A simplified version of a Seneca core (NCC)~\cite{tang2023Seneca} is shown in Figure~\ref{fig:multi_layer_network_everything}. It includes a RISC-V controller of pre- and post-processing events (multiplexing/demultiplexing them from/to neurons), the Axon Message Interface (AMI) for input/output buffering of events, a vector of 8 Neuro-Processing Elements (NPEs) that execute SIMD operations for neurons and other functions in parallel, local SRAM memory for storing neuron and other state variables. Typically, each layer of a neural network model is mapped to one or more NCCs, although if a layer is sufficiently small, then multiple layers can also be mapped onto one NCC. NCCs communicate through a Network-on-Chip (NoC). The RISC-V controller is programmable in C, while NPE are programmable in Assembly-style microcode.

In this architecture and for the testing purposes of this work, the Delay IP has been integrated at every core between the AMI and the NoC (Figure~\ref{fig:multi_layer_network_everything}). A full and more compact integration could make it part of the AMI or part of the NoC since there are queues already in both components, thereby leading to more memory/latency improvements (than what is shown in the results).

\subsection{Experimental Setup}

The models used for the tests in this section undergo initial training and testing in PyTorch.
Next, they are mapped to Seneca (bfloat16 quantization) and Loihi (int8 quantization) for measuring latency, energy consumption, area utilization, and memory usage.

Since Seneca is not taped out yet, we have estimated its power consumption by using power estimations based on a netlist provided by Imec, in a 22 nm process from Global Foundry. Energy and area measurements are acquired using the Cadence Joules and Genus tools, respectively.
The Joules tool provides accuracy within 15\% of the sign-off power.
The Genus tool is used to estimate the area. Latency, memory, and fidelity measurements are obtained from hardware test bench simulations. The same models are run in PyTorch to assert fidelity between hardware and software.

We trained three different SNN models featuring LIF neurons, on the Spiking Heidelberg Digits benchmark~\cite{zenke2022shd} (input size 700, and 20 output classes). To optimize simulation time, we carefully designed a test set comprising 100 unique data points, allowing us to individually assess each class with the networks we examined.

Each model consists of four layers of neurons, with synaptic delays between each pair of layers, except for the first layer (input-hidden1). The structure of the neuron/layer configurations for these three models is as follows: 700-48-48-20, 700-32-32-20, and 700-24-24-20. We set a maximum delay of 60 discrete time steps with a stride of 2, i.e. pruning every second delay axon. Furthermore, we applied \emph{per-synapse} delay pruning during training, based on the approach outlined in~\cite{patino2023empirical}, effectively reducing the number of synapses to only 15 per pair of neurons.

For the 700-48-48-20 model, we also explored a variation involving \emph{axonal-only delay pruning}. This means that instead of removing individual synapses, we removed entire groups of synapses per axon until only 15 delay axons remained (700-48-48-20 Ax). The goal here was to understand the trade-off in hardware resource savings between these two pruning strategies.


For each model on Seneca, we occupied 3 NCCs, one projection per core (as shown in Figure \ref{fig:multi_layer_network_everything}), while on Loihi we could fit each model in 1 core. 
Additionally, in Seneca, tests were done with the Delay IP and without the Delay IP (where the SCDQ was implemented in C on the RISC-V controller).

\section{Results}
\label{sec:results}

\subsection{Software-Hardware Model Fidelity}

\begin{figure}
\centering
\includegraphics[width=0.48\textwidth,trim=5 5 5 5,clip]{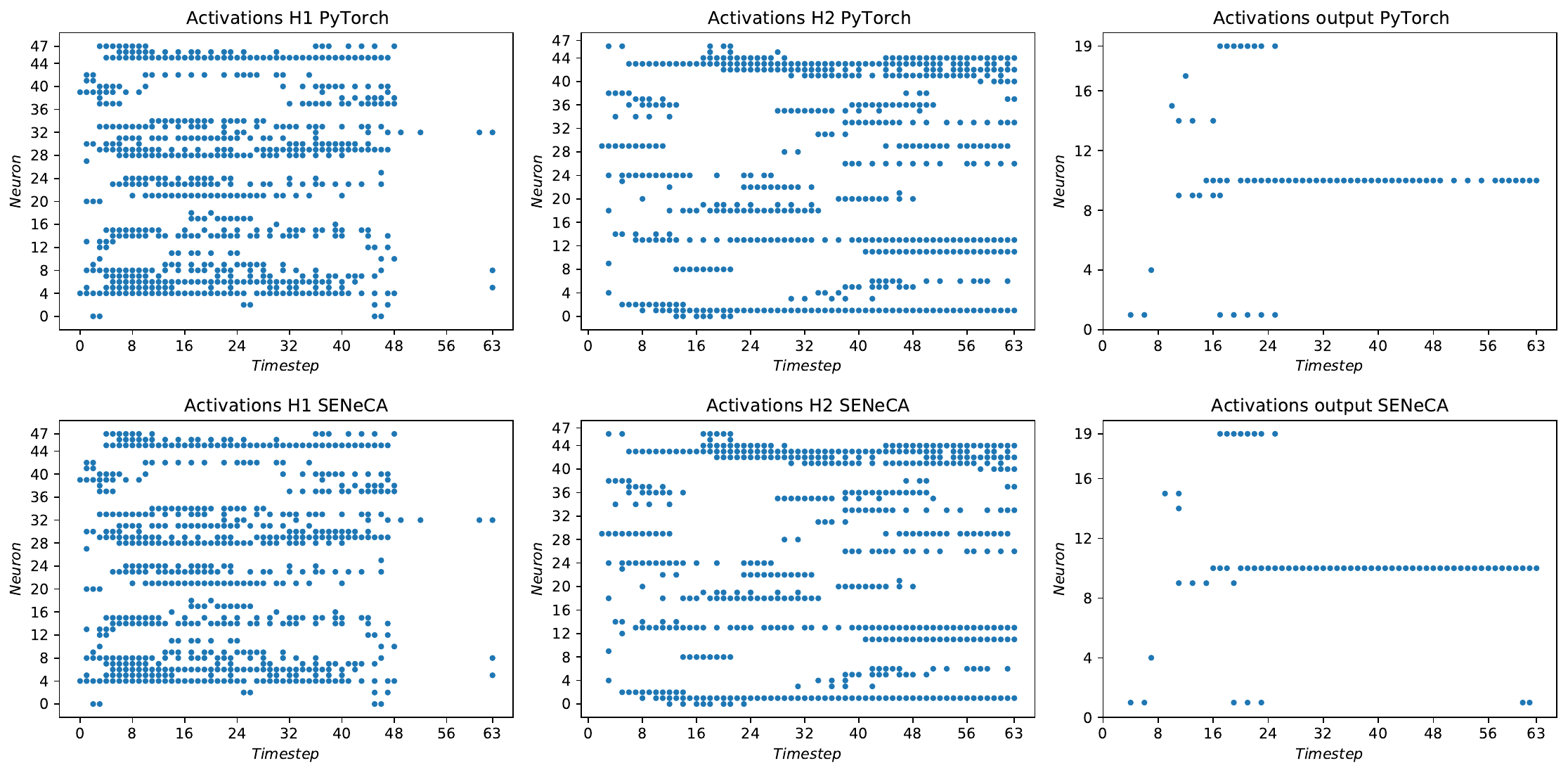}
\caption{\small Activations of a datapoint in all layers for 48-48-20 network on Seneca, Loihi and PyTorch}
\label{fig:activations}
\end{figure}

\begin{figure}
\centering
\includegraphics[width=0.48\textwidth,trim=5 5 5 5,clip]{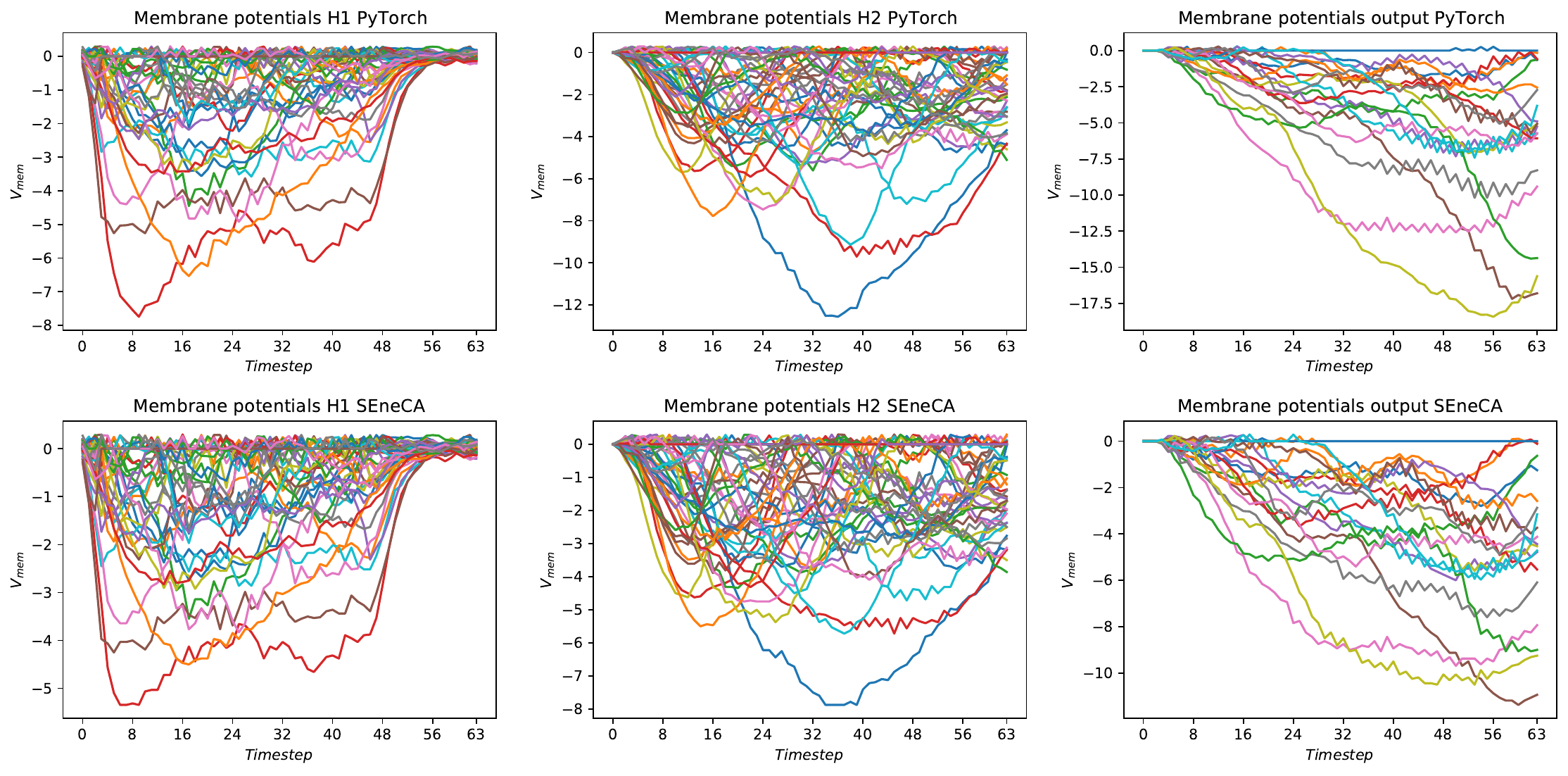}
\caption{\small Membrane potentials of a data point in all layers for 48-48-20 network on Seneca, Loihi and PyTorch}
\label{fig:membranes}
\end{figure}

Table \ref{tb:accuracy_results} reports on the fidelity between the software run model and the hardware model deployed on Seneca. 
The results confirm the consistency between PyTorch, and Seneca. This consistency is further visualized in Figures \ref{fig:activations} and \ref{fig:membranes}, which show for one data point the evolution of the neuron activations and the neuron membrane state. The accuracy of 48-48-20 Ax (axonal delay pruning) is decreased because of the axon grouping constraint that does include in an axon group the 15 smallest weights. However, it will show improvements in energy and latency.

\begin{table}[]\footnotesize
    \centering
    \caption{\small Fidelity of inference between software (PyTorch) and on-hardware (Seneca) synaptic delay models}
    \renewcommand{\arraystretch}{1.1}    
    \label{tb:accuracy_results}
\begin{tabular}{|ccccc|}
\hline
\multicolumn{1}{|c|}{\textbf{Network}}      & \multicolumn{2}{c|}{\textbf{PyTorch}}  & \multicolumn{2}{c|}{\textbf{Seneca}} \\ \hline
\multicolumn{5}{|c|}{Model accuracy}                                                                                                                                              \\ \hline
\multicolumn{1}{|c|}{48-48-20}              & \multicolumn{2}{c|}{87\%}              & \multicolumn{2}{c|}{86\%}            \\
\multicolumn{1}{|c|}{32-32-20}              & \multicolumn{2}{c|}{82\%}              & \multicolumn{2}{c|}{81\%}            \\
\multicolumn{1}{|c|}{24-24-20}              & \multicolumn{2}{c|}{83\%}              & \multicolumn{2}{c|}{82\%}            \\ 
\multicolumn{1}{|c|}{48-48-20 Ax}           & \multicolumn{2}{c|}{76\%}              & \multicolumn{2}{c|}{76\%}            \\ \hline
\multicolumn{5}{|c|}{Consistency against PyTorch}                                                                                                                      \\ \hline
\multicolumn{1}{|c|}{48-48-20}              & \multicolumn{2}{c|}{100\%}             & \multicolumn{2}{c|}{93\%}            \\
\multicolumn{1}{|c|}{32-32-20}              & \multicolumn{2}{c|}{100\%}             & \multicolumn{2}{c|}{97\%}            \\
\multicolumn{1}{|c|}{24-24-20}              & \multicolumn{2}{c|}{100\%}             & \multicolumn{2}{c|}{98\%}            \\
\multicolumn{1}{|c|}{48-48-20 Ax}           & \multicolumn{2}{c|}{100\%}             & \multicolumn{2}{c|}{94\%}            \\ \hline
\end{tabular}
\end{table}

\subsection{Energy, Latency \& Memory}

\begin{table}[] 
    \footnotesize
    \centering
    \caption{\small Avg hardware measurements per inference, DIP = Delay IP, Energy calculated using dynamic power consumption}
    \renewcommand{\arraystretch}{1.2}    
    \label{tb:hw_metrics}
    \begin{tabular}{|c|ccc|}
        \hline
        \textbf{Measurement} & \textbf{Loihi (1c)} & \textbf{Seneca (3c) w/ DIP} & \textbf{Seneca (3c)  w/o DIP} \\
        \hline
        \multicolumn{4}{|c|}{\textbf{Network: 48-48-20}}\\
        \hline 
        \textit{Energy}       & 28.4 $\mu J$     			& 43.6 $\mu J$   	& 152 $\mu J$\\ 
        \textit{Latency}      & 9.1 $ms$					& 4.25 $ms$   		& 18.3 $ms$ 	\\
        \textit{Memory}       & 1.96 $Mb$         			& 3.38 $Mb$   		& 3.34 $Mb$  	\\
        \textit{DIP energy}   & NA             				& 0.98 $\mu J$   	& 0 $\mu J$    	\\
        \hline 
        \multicolumn{4}{|c|}{\textbf{Network: 32-32-20}}										\\
        \hline 
        \textit{Energy}       & 18.3 $\mu J$                & 29.5 $\mu J$  	& 92.4 $\mu J$  	\\ 
        \textit{Latency}      & 9.1 $ms$                    & 3.34 $ms$   		& 13.6 $ms$ 	\\
        \textit{Memory}       & 1.48 $Mb$         	        & 2.33 $Mb$   		& 2.29 $Mb$  	\\
        \textit{DIP energy}   & NA             			    & 0.77 $\mu J$   	& 0 $\mu J$    	\\
        \hline 
        \multicolumn{4}{|c|}{\textbf{Network: 24-24-20}}										\\
        \hline 
        \textit{Energy}       & 12.5 $\mu J$                & 23.8 $\mu J$   	& 73.3 $\mu J$ 	\\ 
        \textit{Latency}      & 9.1 $ms$         	        & 2.81 $ms$   		& 9.81 $ms$ 	\\
        \textit{Memory}       & 1.23 $Mb$         	        & 2.07 $Mb$   		& 2.05 $Mb$  	\\
        \textit{DIP energy}   & NA                          & 0.62 $\mu J$      & 0 $\mu J$    						\\
        \hline 
        \multicolumn{4}{|c|}{\textbf{Network: 48-48-20 Ax}}										\\
        \hline 
        \textit{Energy}       & 25.2 $\mu J$                & 31.4 $\mu J$      & NA 			\\ 
        \textit{Latency}      & 10.0 $ms$         	        & 3.03 $ms$   		& NA 			\\
        \textit{Memory}       & 1.96 $Mb$         	        & 3.36 $Mb$   		& NA 			\\
        \textit{DIP energy}   & NA             			    & 0.72 $\mu J$      & NA 			\\
        \hline 
    \end{tabular} 
\end{table}

Table \ref{tb:hw_metrics} reports (average) hardware measurements per inference in Loihi, in Seneca with Delay IP, and without Delay IP. Energy is calculated using dynamic power consumption, to exclude the high static power consumption of the Latermont of Loihi of 74.4 mW. The static power consumption of Seneca is 462 $\mu W$. Comparing the middle and right columns justifies the circuit-based implementation of SCDQ as it improves energy efficiency and latency by a significant amount. The energy efficiency increases between $3.5 \times$ (48-48-20) and $3.1 \times$ (24-24-20). Similarly, latency decreases between $4.3 \times$ (48-48-20) and $3.5 \times$ (24-24-20). In other words, the benefits increase with the total activations volume (seen in larger networks). For the axonal pruned network (48-48-20 Ax), both energy efficiency and latency improve by approximately $1.4 \times$ compared to the synaptic pruned network (48-48-20). This is also expected since the activation sparsity is higher for 48-48-20 Ax than for 48-48-20.

In comparison to Loihi (Ring Buffer), the energy efficiency of Seneca (SCDQ) appears approximately $1.2-1.9 \times$ lower, but note that this involves 1 core in Loihi versus 3 cores on Seneca.
However, the gap is closing as the network size increases and that is justified by the costly contribution of the RISCV controller (the larger the network, the more compute work can be allocated to the NPEs).

In Seneca Delay IPs contribute only 2\% - 3\% to this energy consumption, with the remaining 97\% - 98\% coming from the NCCs (of which 67\% - 82\% of this energy consumption is related to pre- and post-processing events in RISC-V cores for networks with Delay IPs, and 86\% - 88\% for the networks without Delay IP).

What is more interesting for the role of SCDQ is the inference latency. On Seneca, both with the circuit implementation and the RISC-V implementation of SCDQ, the latency decreases with the network size (due to less activations), while it, surprisingly, remains constant on Loihi's Ring Buffers. And for the circuit implementation (Delay IP), it is consistently less than $0.5 \times$ that of Loihi, despite the larger number of cores used and routing between them. 

\begin{figure}
\centering
\includegraphics[width=0.40\textwidth,trim=5 5 5 5,clip]{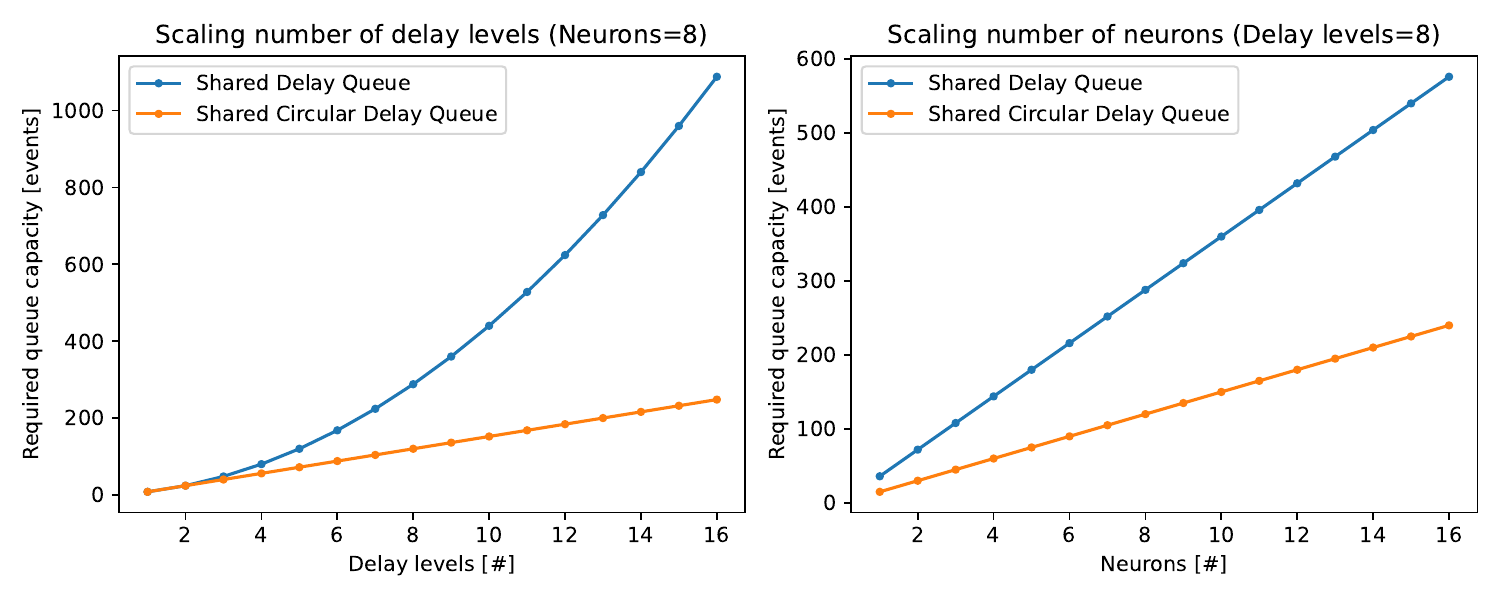}
\caption{\small Required memory for shared delay queue vs shared circular delay queue}
\label{fig:queue_size_scaling}
\end{figure}

In Section \ref{sec:background}, we touched upon the better memory scalability of a Shared Delay Queue compared to the Ring Buffer implementation of synaptic delays. This is also attested in~\cite{patino2023empirical}. To illustrate how SCDQ improves upon the \emph{vanilla} Shared Delay Queue structure~\cite{akopyan2015truenorth}, in Figure~\ref{fig:queue_size_scaling}, we provide a worst-case estimate (assuming dense activations and no delay pruning) between two fully connected layers. The figure illustrates the buffer capacity required for both a Shared Delay Queue and SCDQ, as the number of delay levels increases and as the number of postsynaptic neurons increases. It is evident that SCDQ exhibits superior scalability (next to being also more flexible).



\subsection{Area Optimizations}
\label{subsec:area}

Taking $EDAP = E \cdot L \cdot A$ as a performance metric that combines energy per inference $E$, latency per inference $L$, and area $A$, where lower means better, we find that for the 48-48-20 model
\footnote{One Seneca NCC's area is 472400 $\mu m^2$, and one Delay IP's area is 81624 $\mu m^2$ without area optimizations, resulting in 1.58 $mm^2$ for running the network on Seneca.}, Seneca achieves $43.6 \mu J \cdot 4.25 ms \cdot 1.58 mm^2 \approx 2.93 \cdot 10^{-13} Jsm^2$.

In the initial Delay IP implementation of SCDQ, the memory block in Figure \ref{fig:multi_layer_network_everything} was implemented with flip-flops. To reduce the total area, flip-flop memory was replaced with a 2048-word 16-bit SRAM memory, with enough capacity for the observed maximum of 1596 events in the queues. This resulted in an 81\% reduction in total area, with a total of 15463 $\mu m^2$ for the Delay IP (see Table \ref{tb:area}). In this case $EDAP$ for Seneca  $43.6 \mu J \cdot 4.25 ms \cdot 1.45 mm^2 \approx 2.68 \cdot 10^{-13} Jsm^2$ for the 48-48-20 network. With this area optimization the Delay IP reduces from $\frac{81624}{472400} = 17\%$ of the area one Seneca core down to a mere $\frac{15463}{472400} = 3\%$ only of the size of one core.

\begin{table}[ht]
    \footnotesize
    \centering
    \setlength{\tabcolsep}{2pt} 
    \caption{\small Area Savings with SRAM}
    \label{tb:area}
    \begin{tabular}{|c|cc|cc|cc|c|}
    \hline
   \textbf{Memory in} & \multicolumn{2}{c|}{\textbf{Combinatorial}} & \multicolumn{2}{c|}{\textbf{Flip-Flop}} & \multicolumn{2}{c|}{\textbf{SRAM}} & \textbf{Total}\\\hline
   \textit{FF}   & 16885 $\mu m^2$ & 21\% & 64739 $\mu m^2$ & 79\% & 0 $\mu m^2$ & 0\% & 81624 $\mu m^2$ \\\hline
   \textit{SRAM} & 2287 $\mu m^2$ & 15\% & 4557 $\mu m^2$ & 29\% & 8618 $\mu m^2$ & 56\% & 15463 $\mu m^2$ \\\hline
   \multicolumn{7}{|c|}{\textit{Area saved}} & 81\%\\\hline
\end{tabular}
\end{table}

\vspace{-1em}
\section{Discussion and Future Work}
\label{sec:discussion}

\begin{figure}
\centering
\includegraphics[width=0.48\textwidth,trim=5 5 5 5,clip]{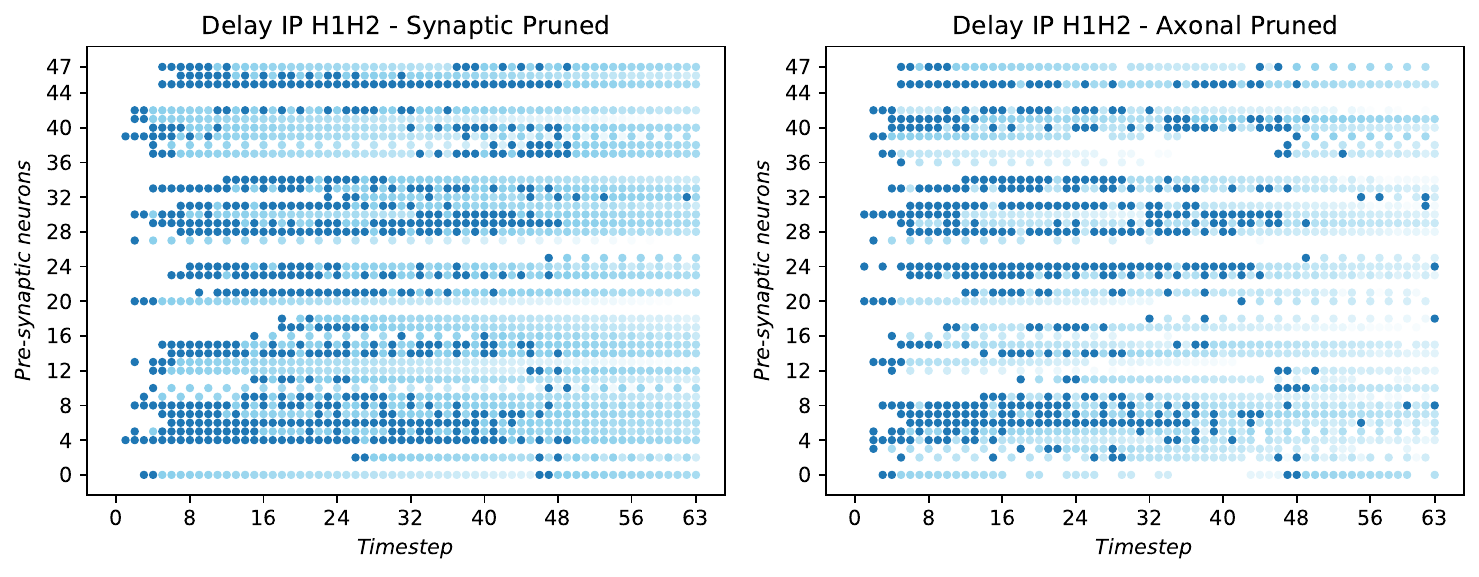}
\caption{\small Event trace within a Delay IP of Seneca, showing for a datapoint the behavior in both the 48-48-20 synaptic pruned (left) and axonal pruned (right) network. Each dark dot represents an event exiting the queue for the first time, while light dots indicate subsequent exits, with lighter shades denoting it is closer to the termination timestep. In the synaptic pruned network, after an event enters the queue, it exits the queue nearly every timestep during inference, and it is terminated at the maximum delay. In the axonal pruned network, an event that enters the queue does not exit nearly every timestep and it is terminated before the maximum delay.}
\label{fig:delay_queue_content}
\end{figure}

Seneca~\cite{tang2023Seneca} as an event-driven processor uses source-routing of event packets (AER) from a presynaptic neuron to postsynaptic neurons. A presynaptic event is thus broadcast to all postsynaptic neurons after exiting the SCDQ. As a result, and due to the way we model synaptic delay levels using separate axons\cite{patino2023empirical}, pruning a single connection to a postsynaptic neuron (synapse) at the model will reduce neither incoming traffic nor outgoing traffic (and memory use) in SCDQ. That is, an event from a presynaptic neuron will remain in the queue for delivery to the remaining synapses at other delay steps. However, if all synapses of an axon are pruned, there is no recipient neuron for an event at a certain delay step. In this case, the event can be dropped earlier from the SCDQ, reducing memory use. For this reason, axonal pruning \emph{may} also reduce latency (and energy). This is confirmed in Table\ref{tb:hw_metrics} and also in Figure \ref{fig:delay_queue_content}, where dark dots represent the times that events are fired from presynaptic neurons and lighter colored dots represent the \emph{lifespan} of these events in the SCDQ thereafter (fading based on the counter values).

In Seneca, it is possible to share a core across network layers, which would mean that more than one layer can in fact share the same Delay IP. If there is no barrier synchronization between layers (EOT events), this is likely to further improve inference latency in an event-driven accelerator and also improve the hardware efficiency of the Delay IP. But to maintain task performance of the model in this case requires a more sophisticated training procedure and algorithm-hardware co-optimizations, which is left for future work.

SCDQ has memory overhead of $\alpha \cdot I \cdot (2\cdot D - 1)$, scaling with $\mathcal{O}(\alpha \cdot I \cdot D)$.
By reference to TrueNorth's\cite{akopyan2015truenorth} constraints of 16 timesteps of delay, 256 neurons, and for sparsity $\alpha = 1$ (no spasity), TrueNorth's Shared Delay Queue needs queue capacity for $34816$ events, while SCDQ only requires storage for $7936$ events.
Each event in the Delay IP instantiation of SCDQ has a bit width of 16 bits. With reference to Loihi's\cite{davies2018loihi} constraints of 64 delay time steps, 48 neurons, and 8 bits per weight, the additional memory overhead of Loihi is only $24576$ bits, which is less than the $97536$ bits of the Delay IP. However, this comparison assumes no activation sparsity ($\alpha = 1$), and the memory overhead of Loihi does not scale with activation sparsity. If $\alpha \leq 0.25$, the SCDQ has a lower memory overhead than Loihi.
SpiNNaker\cite{khan2008spinnaker} has a constraint of 16 time steps of delay, 256 neurons, and 16-bit weights per neuron, resulting in a memory overhead of $65536$ bits for the Ring Buffers. Assuming $\alpha = 1$, the SCDQ has a memory overhead of $126976$ bits in this case and when $\alpha \leq 0.5$, SCDQ is more memory efficient than SpiNNaker.

To further improve the memory efficiency of the SCDQ, the implementation can be modified to use only one FIFO instead of two and to store pointers to shared delay counters in events instead of the delay counter itself. Only the $D$ delay counters need to be maintained in this case. Utilizing a single FIFO can be achieved by writing incoming events to the FIFO and reading them multiple times until they are no longer used. At the end of a timestep, only the read and write pointers need to be modified according to the delay values. This improvement is currently tested in software and seems to improve memory efficiency by approximately 2$\times$, reducing the total memory requirements to $\alpha \cdot I \cdot D$ (previously $\alpha \cdot I \cdot (2\cdot D - 1)$.

\section{Conclusion}


This work introduces SCDQ, a novel hardware structure for enabling synaptic delays on neuromorphic accelerators. It builds upon and improves previous designs and approaches~\cite{akopyan2015truenorth,davies2018loihi,khan2008spinnaker}. It scales better in terms of memory and area, and is more amortizable to algorithm-hardware co-optimizations; in fact, memory scaling is modulated by model sparsity and not merely network size. It has negligible energy overhead and halves inference latency.

\bibliographystyle{unsrt} 
\bibliography{bio}


\end{document}